\documentclass[conference]{IEEEtran}
\IEEEoverridecommandlockouts
\usepackage{cite}
\usepackage{amsmath,amssymb,amsfonts}
\usepackage{algorithmic}
\usepackage{graphicx}
\usepackage{textcomp}
\usepackage{xcolor}
\def\BibTeX{{\rm B\kern-.05em{\sc i\kern-.025em b}\kern-.08em
    T\kern-.1667em\lower.7ex\hbox{E}\kern-.125emX}}

\usepackage[draft]{pgf}
\usepackage{physics}
\usepackage{siunitx}
\usepackage{graphics} 
\usepackage{graphicx} 
\usepackage{epsfig} 
\usepackage{amsmath, bm} 
\usepackage{amssymb}  
\usepackage{upgreek}

\usepackage{tensor}
\usepackage[ruled,vlined]{algorithm2e}

\usepackage{cuted}

\usepackage{booktabs}
\usepackage{multicol}
\usepackage{multirow}
\usepackage{stfloats}
\usepackage{lipsum}  
\usepackage{breqn}
\usepackage{comment} 

\usepackage[caption=false]{subfig} 

\graphicspath{ {./images/} }
\title{\LARGE \bf Unilateral Ground Contact Force Regulations \\in Thruster-Assisted Legged Locomotion
}

\author{\IEEEauthorblockN{Eric Sihite}
\IEEEauthorblockA{\textit{Electrical and Computer Engineering} \\
\textit{Northeastern University}\\
Boston, USA \\
e.sihite@northeastern.edu}
\and
\IEEEauthorblockN{Pravin Dangol}
\IEEEauthorblockA{\textit{Electrical and Computer Engineering} \\
\textit{Northeastern University}\\
Boston, USA \\
dangol.p@northeastern.edu}
\and
\IEEEauthorblockN{Alireza Ramezani}
\IEEEauthorblockA{\textit{Electrical and Computer Engineering} \\
\textit{Northeastern University}\\
Boston, USA \\
a.ramezani@northeastern.edu}
}

\begin{document}

\maketitle
\thispagestyle{empty}
\pagestyle{empty}

\begin{abstract}

In this paper, we study the regulation of the Ground Contact Forces (GRF) in thruster-assisted legged locomotion. We will employ Reference Governors (RGs) for enforcing GRF constraints in \textit{Harpy} model which is a bipedal robot that is being developed at Northeastern University. Optimization-based methods and whole body control are widely used for enforcing the no-slip constraints in legged locomotion which can be very computationally expensive. In contrast, RGs can enforce these constraints by manipulating joint reference trajectories using Lyapunov stability arguments which can be computed much faster. The addition of the thrusters in our model allows to manipulate the gait parameters and the GRF without sacrificing the locomotion stability.

\end{abstract}



\section{Introduction}

There are several examples of successful legged robots that can hop or trot robustly in the presence of significant disturbances, such as the Raibert's hopping robots \cite{raibert1984experiments} and Boston Dynamics' robots \cite{raibert2008bigdog}. Other than these successful examples, a large number of underactuated and fully actuated bipedal robots have also been introduced. Agility Robotics' Cassie \cite{gong2019feedback}, Honda's ASIMO \cite{hirose2006honda} and Samsung's Mahru III \cite{kwon2007biped} are capable of walking, running, dancing and going up and down stairs, and the Yobotics-IHMC \cite{5354430} biped can recover from pushes. Despite these accomplishments, these systems are prone to falling over when navigating rough terrains. Even humans, which has naturally robust gait, can trip and fall over when walking on uneven or slippery surfaces. Therefore, the objective of this work is to extend our knowledge on bipedal walking and explore the possibility of using thrusters to assist bipedal robots to achieve a more stable walking gait. 

\begin{figure}[t]
    \vspace{0.05in}
    \centering
    \includegraphics[width = 0.7 \linewidth]{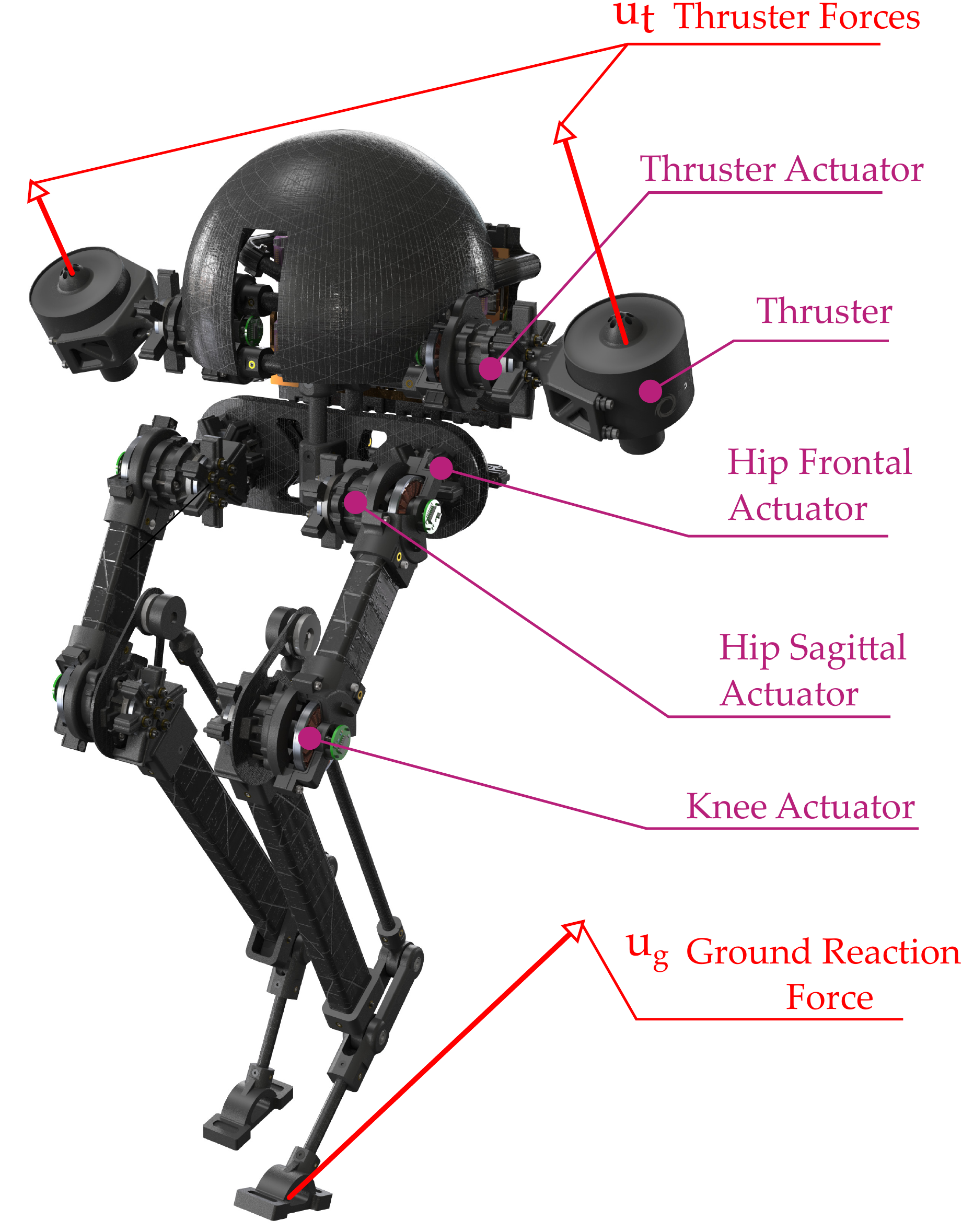}
    \caption{Illustration of a concept design for \textit{Harpy}, a thruster-assisted bipedal robot designed by the authors to study robust, efficient and agile legged robotics.}
    \label{fig:midget}
    \vspace{-0.05in}
\end{figure}

In this paper, we will report our efforts in designing closed-loop feedback for the thruster-assisted walking of a legged system called \textit{Harpy} (shown in Fig.~\ref{fig:midget}), currently its hardware being developed at Northeastern University. This biped is equipped with a total of eight actuators, and a pair of coaxial thrusters fixed to its torso. The thrusters allow the robot to perform multi-modal locomotion, where it can simply fly over difficult terrains where walking can be highly costly or difficult for the robot to handle. 

Thrusters can result in unparalleled capabilities. For instance, gait trajectory planning (or re-planning), control and unilateral contact force regulation can be treated significantly differently as we have shown previously \cite{dangol2020performance,de2020thruster,dangol2020towards,dangol2020feedback,liang2021rough}. That said, real-time gait trajectory design in legged robots has been widely studied and the application of optimization-based methods is very common \cite{hereid20163d}. The optimization allows the implementation of constraints to avoid slipping, but these methods can be cumbersome as they are widely defined based on Whole Body Control (WBC) which can lead to computationally expensive algorithms \cite{sentis2006whole}. Other attempts entail optimization-based, nonlinear approaches to secure safety and performance of legged locomotion \cite{CLFQP,7803333,7041347}. 

We will capitalize on the thrusters action in Harpy and will show that one can limit the use of costly optimization-based schemes by directly regulating contact forces. We will resolve gait parameters and re-plan them during the whole Single Support (SS) phase, which is the longest phase in a gait cycle, by only assuming well-tuned supervisory controllers found in \cite{sontag1983lyapunov, 371031, bhat1998continuous} and by focusing on fine-tuning the joints desired trajectories to satisfy unilateral contact force constraints. To do this, we will devise intermediary filters based on the celebrated idea of Explicit Reference Governors (ERG) \cite{411031,bemporad1998reference,gilbert2002nonlinear}. ERGs relied on provable Lyapunov stability properties can perform the motion planning problem in the state space in a much faster way than widely used optimization-based methods. That said, these ERG-based gait modifications and impact events (i.e., impulsive effects) can lead to severe deviations from the desired periodic orbits and standard legged robots cannot sustain these perturbations. Previously, we demonstrated that the thrusters can be leveraged to enforce hybrid invariance in a robust fashion by applying predictive schemes within the Double Support (DS) phase \cite{dangol2020towards}.


In this paper, we explore the implementation of ERG in enforcing ground reaction force (GRF) constraints on a bipedal robot. First, the dynamic and reduced order models of the robot are derived where the addition of thrusters allow a fully actuated variable length inverted pendulum (VLIP) model. The VLIP model will be used to model the GRF which will be used to calculate the no-slip constraints and be enforced by the ERG. The implementation of the ERG is done on the VLIP model and on the 3D biped model where we will show the performance of applying the ERG on these systems. This paper is outlined as follows: the dynamic modeling for Harpy reduced-order model (ROM) which will be used in designing the ERG, the ERG algorithm used in this paper, and followed by the numerical simulations and the concluding remarks.

\section{Dynamic Modeling and Control}

This section contains the brief overview of the dynamic model used in this paper for the simulation, which is followed with the derivation of the ROM to be used in the ERG and controller design.  

\label{subsec:modeling_full}

\begin{figure}[t]
    \centering
    \includegraphics[width=0.7\linewidth]{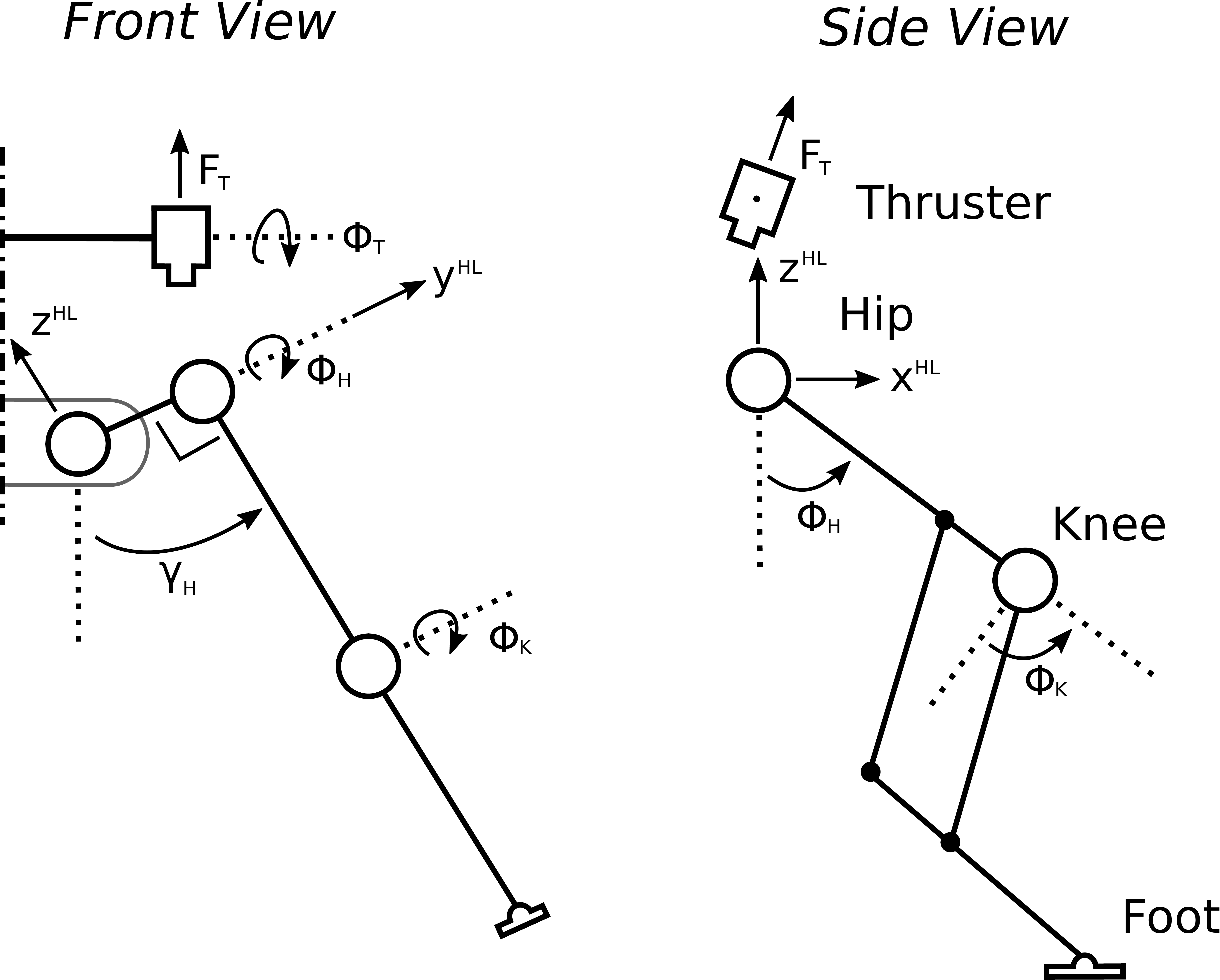}
    \caption{Leg kinematics of the robot where the leg joints are represented by the hip frontal ($\gamma_H$), hip sagittal ($\phi_H$), and knee sagittal ($\phi_K$) angles. The thrusters are designed to rotate about the sagittal angle; however in this paper, in order to simplify the controls, they can apply force in any direction.}
    \label{fig:harpyLegKinematics}
\end{figure}

Figure \ref{fig:harpyLegKinematics} shows the degrees of freedom of the robot's leg where there are three actuated joints: hip frontal, hip sagittal, and knee sagittal joints. Combined with the robot's body, the system has a combined total of 12 degrees-of-freedom (DoF). The thrusters are designed to rotate about the body's sagittal axis, but in the current modeling we assume that the thrusters can provide forces in any direction to simplify the problem. In this case, the thruster dynamics is also ignored. The model is simplified further by assuming that the mass is concentrated at the body and the joints motors, which results in a simpler model where the lower leg (shin and foot) are massless. The foot is also considered to be small so they can be modeled as a point foot which simplifies the ground force effect to the system at the cost of less stability due to the smaller support polygon.

The dynamic model of Harpy, which is used in the numerical simulation, can be derived using the Euler-Lagrangian dynamic formulation. The body rotation is derived using the modified Lagrangian for dynamics in SO(3) which is done to avoid the gimbal lock or singularity which exists in the Tait-Bryan representation of the rotation matrix. Let $\bm x$ be the system states, defined as follows
\begin{equation}
    \bm x = [\bm c; \bm r_B; \bm \gamma_H; \bm \phi_H; \bm \phi_K;
            \dot{\bm c}; \bm \omega_B; \dot{\bm \gamma}_H; \dot{\bm \phi}_H; \dot{\bm \phi}_K],
    \label{eq:full_states}
\end{equation}
where $\bm c$ is the inertial position of the body center of mass, $\bm r_B$ is the vector forming the components of the rotation matrix $\bm z = \bm R_B\, \bm z^B$ which rotates from the body frame to the inertial frame, $\bm \omega_B$ is the body angular velocity about the body frame. Furthermore, $\bm \gamma_H$, $\bm \phi_H$, and $\bm \phi_K$ are the vectors representing the leg joint angles (hip frontal, hip and knee sagittal, respectively), where each variables contains the left and right component of the leg joints. Then the system equation of motion can be derived in the standard ODE form
\begin{equation}
    \dot{\bm x} = \bm f (\bm x, \bm u_j, \bm u_t, \bm u_g),
    \label{eq:full_dynamics}
\end{equation}
where $\bm u_j$ is the leg joint actuation inputs, $\bm u_t$ is the thruster forces, and $\bm u_g$ is the GRF. Each of these inputs are separated into left and right leg components (e.g. $\bm u_g = [\bm u_{g,l}; \bm u_{g,r}]$). 

The ground is modeled using the compliant ground model using a very stiff unilateral spring and damping 
\begin{equation}
    u_{gz}(p_{z},\dot p_{z}) = \begin{cases} 
    0 & \mbox{if } p_z > 0, \\
    - k_{pg}\,p_{z} - k_{dg} \dot p_{z} & \mbox{if } p_{z} \leq 0,
    \end{cases}
\end{equation}
where $k_{pg}$ and $k_{dg}$ are the ground spring and damping coefficient respectively, and $p_{z}$ are the foot vertical position. Additionally, $k_{dg} = 0$ when $\dot p_{z} > 0$ which is done to simulate a ground model with undamped rebound. The ground friction forces in the $x$ direction is modeled using the Stribeck friction model
\begin{equation}
    u_{gx}(\dot p_{x}) = \left( -\mu_c + (\mu_s-\mu_c)e^{-(\dot p_x/\sigma)^2}\right)\,u_{gz}\,|\dot p_x| + \mu_v\, \dot p_x,
\end{equation}
where $\mu_s$, $\mu_c$, and $\mu_v$ are the static, Coulomb, and viscous friction coefficients respectively, $\dot p_x$ is the foot velocity in $x$ direction, and $\sigma$ is the Stribeck velocity. The friction forces in $y$ direction can be derived in the same way. Then the GRF can be formed by calculating the $u_{gx}$, $u_{gy}$, and $u_{gz}$ for each leg. The controller design for $\bm u_j$ and $\bm u_t$ will be discussed in Section \ref{subsec:modeling_control}.

\subsection{Reduced-Order Model (ROM)}
\label{subsec:modeling_rom}

\begin{figure}[t]
    \centering
    \includegraphics[width=0.4\linewidth]{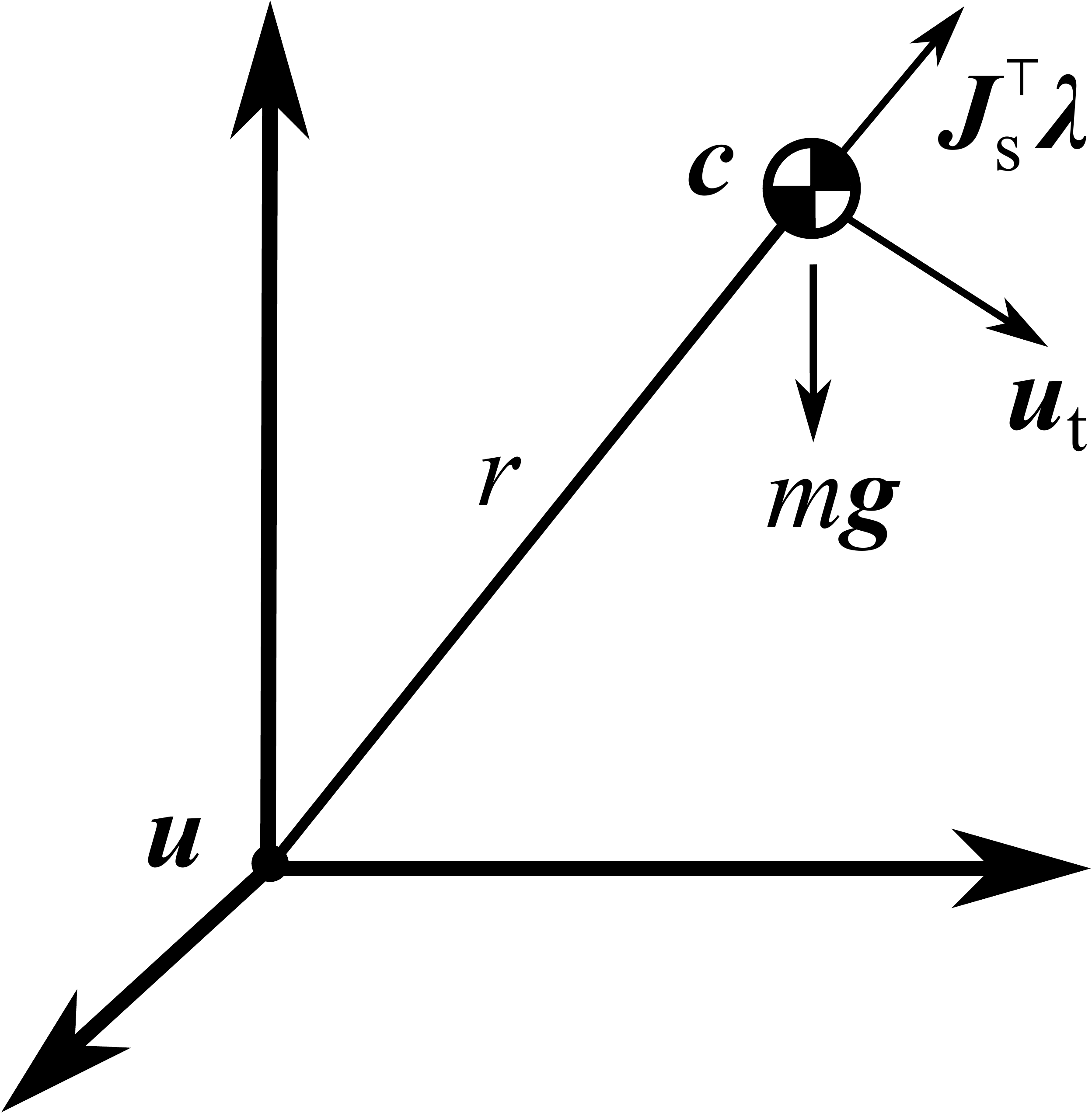}
    \caption{Reduced-order, variable-length, inverted pendulum model with a point-mass is subject to thruster forces. This model will be used to estimate the GRF which will be used to calculate the no-slip constraint equations for the ERG.}
    \label{fig:harpyPendulum}
\end{figure}

The controller for the thruster forces can be designed using the ROM represented by the forced inverted pendulum shown in Fig. \ref{fig:harpyPendulum}. This will be the model used to derive the thruster forces components and the ERG. The dynamic model is derived as follows
\begin{equation}
    \begin{aligned}
    m\,\ddot{\bm c} = m \, \bm g + \bm u_{t,c} + \bm J_s^\top\, \bm \lambda,
    \end{aligned}
\label{eq:harpy_rom}
\end{equation}
where $m$ is the body mass, $\bm g = [0,0,-g]^\top$ is the gravitational acceleration vector, $\bm u_{t,c}$ is the thruster forces about the center of mass, $\bm J_s^\top\, \bm \lambda$ forms the constraint force acting on the body which represents the GRF. The following kinematic constraint equation is implemented
\begin{equation}
    \begin{gathered}
    \bm J_s \,(\ddot{\bm c} - \ddot{\bm u}) = u_r,  \qquad
    \bm J_s = (\bm c - \bm u)^\top, 
    \end{gathered}
\label{eq:harpy_rom_constraint}
\end{equation}
where $\bm u$ is the center of pressure and $u_r$ is the acceleration of the pendulum length. Therefore, $\bm \lambda$ is the Lagrangian multiplier where the constraint equation in \eqref{eq:harpy_rom_constraint} is satisfied and the dynamic equation can be formulated as follows
\begin{equation}
    \begin{bmatrix}
        m \, \bm I & -\bm J_s^\top \\
        \bm J_s & \bm 0
    \end{bmatrix} 
    \begin{pmatrix}
        \ddot{\bm c} \\ \bm \lambda
    \end{pmatrix} =
    \begin{pmatrix}
        m \, \bm g + \bm u_{t,c} \\
        u_r + \bm J_s\,\ddot{\bm u}
    \end{pmatrix}.
\label{eq:harpy_rom_summary}
\end{equation}
Then the Lagrangian multiplier can be solved as follows
\begin{equation}
    \bm \lambda = \left( \tfrac{\bm J_s \bm J_s^\top}{m} \right)^{-1}\,
    \left( -\bm J_s  \left(\bm g + \tfrac{\bm u_{t,c}}{ m}   \right) +  u_r + \bm J_s\,\ddot{\bm u} \right),
\label{eq:harpy_rom_lambda}
\end{equation}
which is used to formulate the GRF
\begin{equation}
    \bm u_g = \bm J_s^\top\,\bm \lambda.
\label{eq:ground_forces}    
\end{equation}
The center of pressure is assumed to be constant ($\ddot{\bm u} = 0$) during the SS phase.

\subsection{Controller Design}
\label{subsec:modeling_control}

The joint controller is designed to track the desired foot positions by using the inverse kinematics to calculate the target joint angles. Let $\bm q = [\bm \gamma_H;\bm \phi_H;\bm \phi_K]$ be the joint angles of the legs. Given the trajectory $\bm q_t$, the joint controller $\bm u_j$ can be derived using the simple PID controller. The trajectory to track is developed by using optimization on a 2D version of the dynamic model shown in \eqref{eq:full_dynamics}. This trajectory is not stable when utilized in the full 3D system which motivates the use of thrusters to stabilize the dynamics.

The feedback control law for both $\bm u_{t,c}$ and $u_r$ are defined as follows
\begin{equation}
    \begin{aligned}
    \bm u_{t,c} &= (\bm I - \bm Y)(\bm K_{pt}(\bm c_t - \bm c) + \bm K_{dt}(\dot{\bm c_t} - \dot{\bm c})) \\
    u_r &= \bm J_s^\top (k_{pr}(\bm c_t - \bm c) + k_{dr}(\dot{\bm c_t} - \dot{\bm c})) \\
    \bm Y &= \bm J_s(\bm J_s^\top \bm J_s)^{-1}\bm J_s^\top,
    \end{aligned}
\label{eq:thruster_feedback_law}
\end{equation}
where $\bm K$ and $k$ are the controller gains, $\bm c_t$ and $\dot{\bm c}_t$ are the trajectories for the center of mass and its velocity. $\bm Y$ is selected to cancel out the radial component of the thruster force along the direction of $\bm c - \bm u$ which reduces the effect of the thruster force to the GRF.

The thruster forces are also separated into the left and right side components ($\bm u_{t,l}$ and $\bm u_{t,r}$ respectively) which is also utilized to stabilize the roll and yaw as follows
\begin{equation}
    \bm u_{t,l} = \begin{bmatrix}
    u_{yaw} \\
    0 \\
    u_{roll} \\
    \end{bmatrix}, \qquad
    \bm u_{t,r} = \begin{bmatrix}
    -u_{yaw} \\
    0 \\
    -u_{roll} \\
    \end{bmatrix},
\label{eq:thruster_controller_LR}    
\end{equation}
\begin{equation}
    \bm u_t = \begin{bmatrix}
    \bm u_{t,c}/2 + \bm u_{t,l} \\
    \bm u_{t,c}/2 + \bm u_{t,r}
    \end{bmatrix}
\label{eq:thruster_controller_full}
\end{equation}
where $u_{roll}$ and $u_{pitch}$ is the PD controller action to stabilize the body's roll and pitch orientation. The orientation stabilization thruster forces have a net force of zero, which does not affect the reduced-order thruster force used in \eqref{eq:harpy_rom_summary}.

\section{Explicit Reference Governor (ERG) and Enforcing GRF Constraints}
\label{sec:erg}

\begin{figure}[t]
    \centering
    \includegraphics[width=0.75\linewidth]{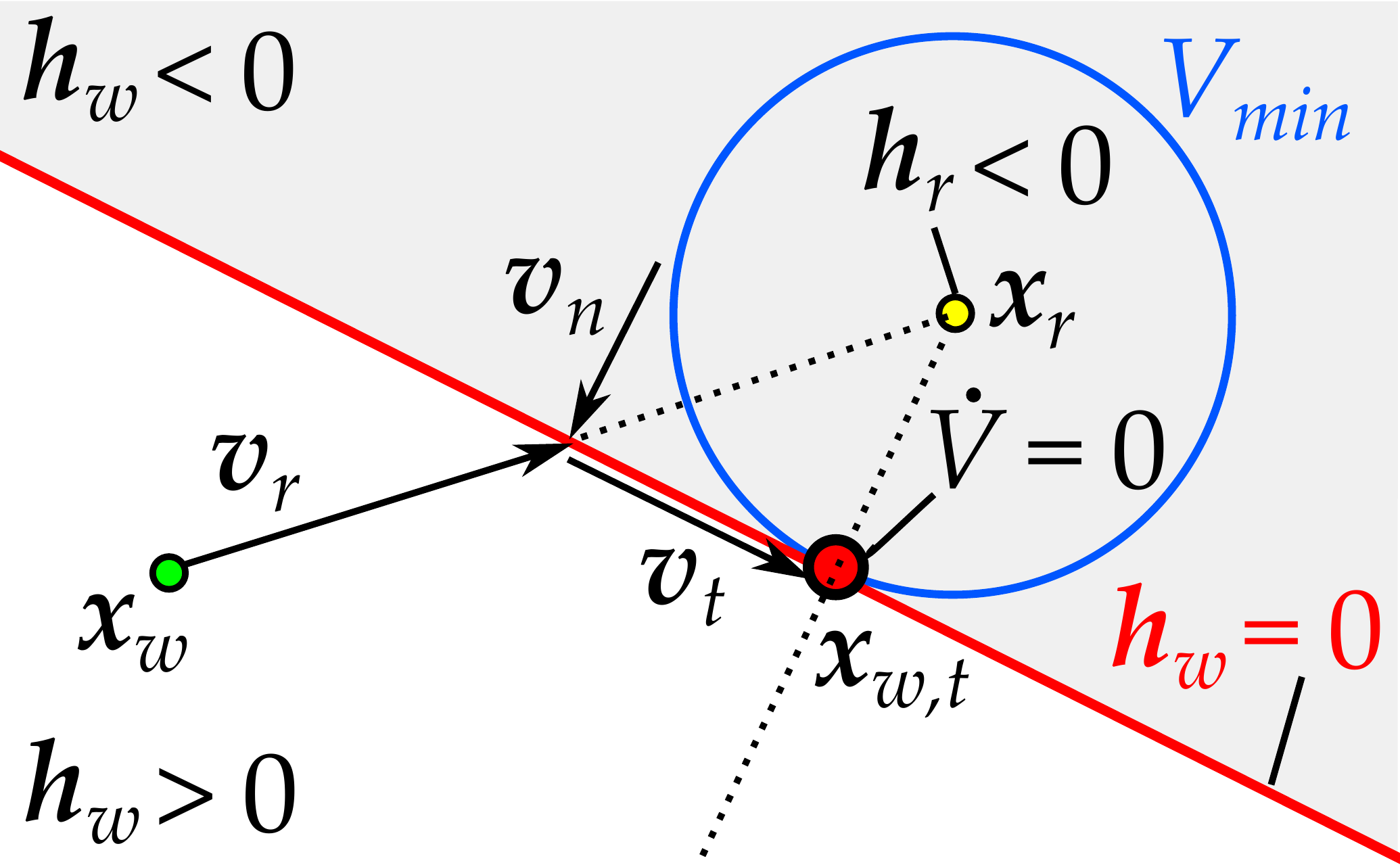}
    \caption{ERG update law for the applied reference to achieve convergence into the minimum energy level set which satisfies the constraint equation defined in \eqref{eq:constraint_eq}. $\bm v_n$ is only applied if the constraint is violated to push the reference into the $\bm h_w > 0$ region.}
    \label{fig:erg_illustration}
\end{figure}

\begin{algorithm}[t]
    \SetAlgoLined
    $\bm{h}_r = \bm{h}_r(\bm{x},\bm{x}_r)$\\
    $\bm{h}_w = \bm{h}_r(\bm{x},\bm{x}_w)$\\
    $\bm v_r = \bm v_t = \bm v_n = \bm 0$\\
    $\bm{C}_r = [ \,\,\,\, ]$\\
    
    \texttt{\\}
    \If{$\min(\bm{h}_w) \geq 0$ \textbf{or} $\min(\bm{h}_r) \geq 0$}{
        $\bm{v}_r = \alpha_r\, (\bm{x}_r - \bm{x}_w)$\\
    }
    \texttt{\\}
    \If{$\min(\bm{h}_w) \geq 0$ \textbf{and} $\min(\bm{h}_r) < 0$}{
        $n_c = \mathrm{length}(\bm h_r)$\\
        \For{$k = 1:n_c$}{
            \If{$h_{r,i} < 0$}{
                $\bm{C}_r = [\bm{C}_r; \bm{J}_r(k,:)]$\\
            }
        }
        $\bm{N}_r = \mathrm{null}(\bm{C}_r)$\\
        $[\sim,n] = \mathrm{size}(\bm{N}_r)$\\
        $\bm{v}_t = \bm 0$\\
        \For{$k = 1:n$}{
            $\bm{n}_k = \bm{N}_r(:,k) / |\bm{N}_r(:,k)|$\\
            $\bm{v}_t = \bm{v}_t + \alpha_t\,\bm{n}_k\, \bm{n}_k^\top\, (\bm{x}_r - \bm{x}_w) $\\
        }
    }
    \texttt{\\}
    \If{$\min(\bm{h}_w) < 0$ \textbf{and} $\min(\bm{h}_r) < 0$}{
       $k_{min} = \min\limits_{k} \bm{h}_w$ (index of the smallest $\bm h_w$)\\
       $\bm r_k = \bm J_r(k_{min},:) / |\bm J_r(k_{min},:)|$\\
        \eIf{$\bm h_r(k_{min}) \geq \bm h_w(k_{min})$}{
            $\bm{v}_n = \alpha_n\,\bm r_k\,r_k^\top\,(\bm x_r - \bm x_w)$\\
        }{
            $\bm{v}_n = -\alpha_n\,\bm r_k\,r_k^\top\,(\bm x_r - \bm x_w)$\\
        } 
    }
    \texttt{\\}
    $\dot{\bm{x}}_w = \bm{v}_r + \bm{v}_t + \bm{v}_n$\\
    $\bm{x}_w = \bm{x}_w + \Delta t\, \dot{\bm{x}}_w$\\
\caption{ERG algorithm}
\label{alg:modified_erg}
\end{algorithm}

The ERG algorithm works by manipulating the controller state reference values such that they are as close as possible to the desired reference trajectory while obeying a set of constraints \cite{gilbert2002nonlinear,garone2015explicit}. The work done in \cite{garone2015explicit} uses a bounded Lyapunov function to show stability and how the constraints are always satisfied by manipulating the reference such that the resulting Lyapunov function is always contained within this boundary. Our version of ERG does not use a Lyapunov function in the manipulated reference update law. Instead, we use a simple heuristic approach where we manipulated the state reference by only using the constraint equation and the system dynamics.

 We assume that the system is controllable and we can track the state reference $\bm{x}_r$. In this ERG formulation, we consider the constraint equations derived in the following form
\begin{equation}
\begin{gathered}
    \bm{h}_r(\bm{x},\bm{x}_r) = \bm{J}_r(\bm x)\, \bm{x}_r + \bm{d}_r(\bm x) \geq 0,
\end{gathered}
\label{eq:constraint_eq}
\end{equation}
which is affine in $\bm{x}_r$. However, some constraints (e.g. ground friction constraints) can't be derived in this form due to the nonlinear nature of the system dynamics. Therefore, an approximation of the constraint equations using Taylor series expansion about $\bm{x}_r$ can be utilized as follows
\begin{equation}
\begin{gathered}
    \bm{J}_r = \left. \left( \tfrac{\partial \bm{h}_r}{\partial \bm{x}_r} \right) \right|_{\bm{x}_r = \bm{x}_{r0}}, \qquad
    \bm{d}_r = \bm{h}_r - \bm{J}_r\, \bm{x}_{r0},
\end{gathered}
\label{eq:taylor_expansion}
\end{equation}
where $\bm{x}_{r0}$ is the current reference value. Since the Jacobian $\bm{J}_r$ forms the rowspace of $\bm{h}_r$ with respect to $\bm{x}_r$, any adjustment in $\bm x_r$ done about the nullspace of $\bm{J}_r$ does not affect $\bm h_r$ which will be utilized in the ERG algorithm to allow partial tracking of $\bm x_r$ when the constraint $\bm h_r$ is violated.

The ERG algorithm can be represented using the applied reference $\bm{x}_w$ which is used in the controller instead of $\bm{x}_r$. The algorithm determines the rate of change of $\bm{x}_w$ such that it's as close as possible to $\bm{x}_r$ while obeying the specified constraints $\bm{h}_w = \bm{h}_r(\bm{x},\bm{x}_w) \geq 0$. Assume that the controller can perfectly track the applied reference $\bm x_w$, i.e. $\bm x = \bm x_w$, then the ERG algorithm can be formulated using the Lyapunov function
\begin{equation}
\begin{gathered}
    V = (\bm x_r - \bm x_w)^\top\, \bm P\, (\bm x_r - \bm x_w),
\end{gathered}
\label{eq:lyapunov_base}
\end{equation}
where $\bm P > 0$ is diagonal, and by assuming that $\dot{\bm x}_r = 0$ then $\dot{V} = 2(\bm x_r - \bm x_w)^\top\, \bm P\, (-\dot{\bm x}_w)$. Then we can select $\dot{\bm x}_w$ such that $\dot V = 0$ at the minimum level set of $V$ that fulfills the constraint $\bm h_w \geq 0$ (defined as $\bm x_{w,t}$), and $\dot V < 0$ if $\min(\bm h_w) \geq 0$ and $\bm x_w \neq \bm x_{w,t}$. Here, $\bm x_{w,t}$ is the closest reference to $\bm x_r$ that satisfy the constraint $\bm h_w \geq 0$, as illustrated in Fig. \ref{fig:erg_illustration}.

Consider the following update law as illustrated in Fig. \ref{fig:erg_illustration} and outlined in Algorithm \ref{alg:modified_erg}
\begin{equation}
\begin{gathered}
    \dot{\bm x}_w = \bm{v}_r + \bm{v}_t + \bm v_n.
\end{gathered}
\label{eq:wd_update_law}
\end{equation}
$\bm v_r$ represents the rate convergence of $\bm x_w$ directly to $\bm x_r$ defined as follows
\begin{equation}
\begin{gathered}
    \bm v_r = \hat \alpha_r\, (\bm x_r - \bm x_w),\\
    \hat \alpha_r = 
    \begin{cases}
     \alpha_r, & \text{if } \min(\bm h_w) \geq 0 \text{ or } \min(\bm h_r) \geq 0 \\
     0,      & \text{else} \\
    \end{cases}
\end{gathered}
\label{eq:vr_derivation}
\end{equation}
where $\alpha_r > 0$. $\bm v_r$ is zero if both the applied and target reference violate the constraints. $\bm v_t$ represents the rate convergence along the nullspace of $\bm J_r$ which allows $\bm x_w$ to partially track $\bm x_r$ because the update about the nullspace does not locally change the value of the constraint equations. Let $\bm C_r$ be the rowspace of the violated constraints of $\bm{h}_r$, and $\bm N_r = \mathrm{null}(\bm C_r) = [\bm{n}_1, \dots, \bm{n}_{n}]$ where $n$ is the size of the nullspace. Let $\bm v_t$ updates $\dot{\bm x_w}$ in the directions of the nullspace as follows
\begin{equation}
\begin{gathered}
    \bm v_t = \textstyle \sum^n_{k=1} \hat{\alpha}_t\, \bm{n}_k\,\bm{n}_k^\top (\bm{x}_r - \bm{x}_w),\\
    \hat \alpha_t = 
    \begin{cases}
     \alpha_t, & \text{if } \min(\bm h_w) \geq 0 \text{ or } \min(\bm h_r) < 0 \\
     0,      & \text{else} \\
    \end{cases}
\end{gathered}
\label{eq:vn_derivation}
\end{equation}
where $\alpha_t > 0$. This update represents the sum of the projections of $(\bm{x}_r - \bm{x}_w)$ into $\bm n_k$. Finally, when the constraints for both $\bm h_w$ and $\bm h_r$ are violated, which might happen if there is a sudden change in parameters (e.g. change in center of pressure $\bm u$), to allow $\bm x_w$ to shift towards the positive constraint values. Let $k$ be the index of the smallest element of $\bm h_w$, and let $\bm r_k$ be the $k$'th row of $\bm J_r$. Then the update towards positive constraint can be derived as follows
\begin{equation}
\begin{gathered}
    \bm v_n = \hat \alpha_n\,\bm r_k\,\bm r_k^\top\, (\bm x_r - \bm x_w) \\
    \hat \alpha_n = 
    \begin{cases}
     \alpha_n, & \text{if } \min(\bm h_w) \leq \min(\bm h_r) < 0 \\
     -\alpha_n, & \text{if } \min(\bm h_r) < \min(\bm h_w) < 0 \\
     0,      & \text{else} \\
    \end{cases}
\end{gathered}
\label{eq:vt_derivation}
\end{equation}
where $\alpha_n > 0$. 

Using the update law defined from \eqref{eq:wd_update_law} to \eqref{eq:vt_derivation} results in
\begin{equation}
\begin{aligned}
    \dot{V} &= - 2(\bm x_r - \bm x_w)^\top\, \bm  Q\, (\bm x_r - \bm x_w), \\
    \bm Q &= \bm P ( \hat \alpha_r\, \bm I + \textstyle \sum^n_{k=1} \hat \alpha_t\, \bm{n}_k \,\bm{n}_k^\top + \hat \alpha_n\,\bm r_k\,\bm r_k^\top).
\end{aligned}
\label{eq:lyapunov_dt_subs}
\end{equation}
$\dot{V} = 0$ if $\min(\bm h_w) \leq 0$ and $\bm n_k \bot (\bm x_r - \bm x_w)$, while $\dot{V} < 0$ when $\min(\bm h_r) \geq 0$ or when $\min(\bm h_w) \geq 0$. This allows the $\bm x_w$ to converge to $\bm x_{w,t}$ which is the minimum energy solution that satisfies $\bm{h}_w \geq 0$ as illustrated in Fig. \ref{fig:erg_illustration}. In case both applied reference and target constraints equation are violated, we have $\dot V > 0$ which drives the $\bm x_w$ towards positive constraint value, away from $\bm x_r$, if $\min(\bm h_r) < \min(\bm h_w)$.

\section{Simulation Results}

\begin{figure}[t]
    \centering
    \includegraphics[width=1\linewidth]{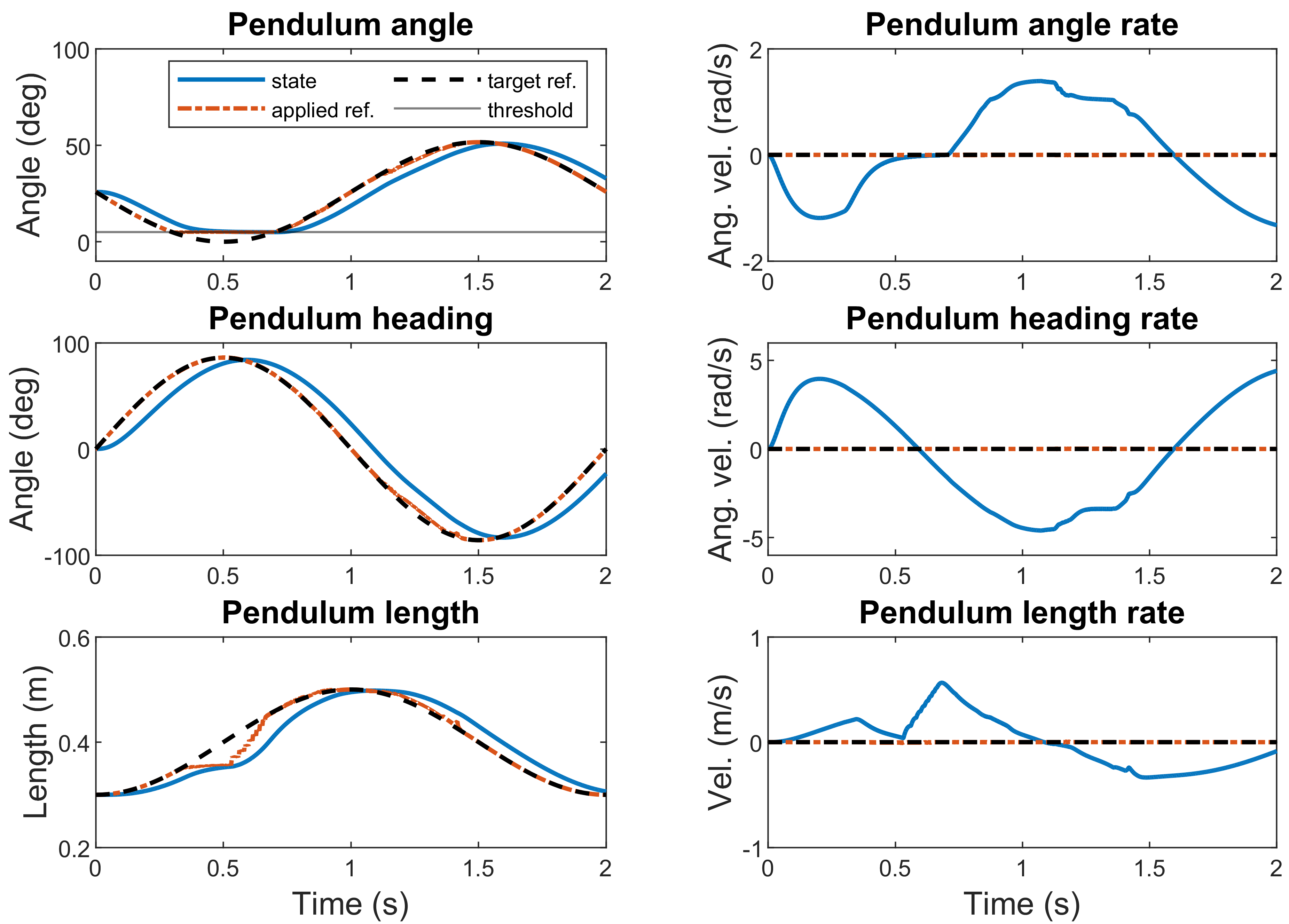}
    \caption{The simulation result of applying ERG on the VLIP model. The ERG manipulated the applied reference trajectory to satisfy the constraints, which can be seen more clearly in the pendulum angle and length around the 0.5 s simulation time.}
    \label{fig:plot_vlip_states}
\end{figure}

\begin{figure}[t]
    \centering
    \includegraphics[width=1\linewidth]{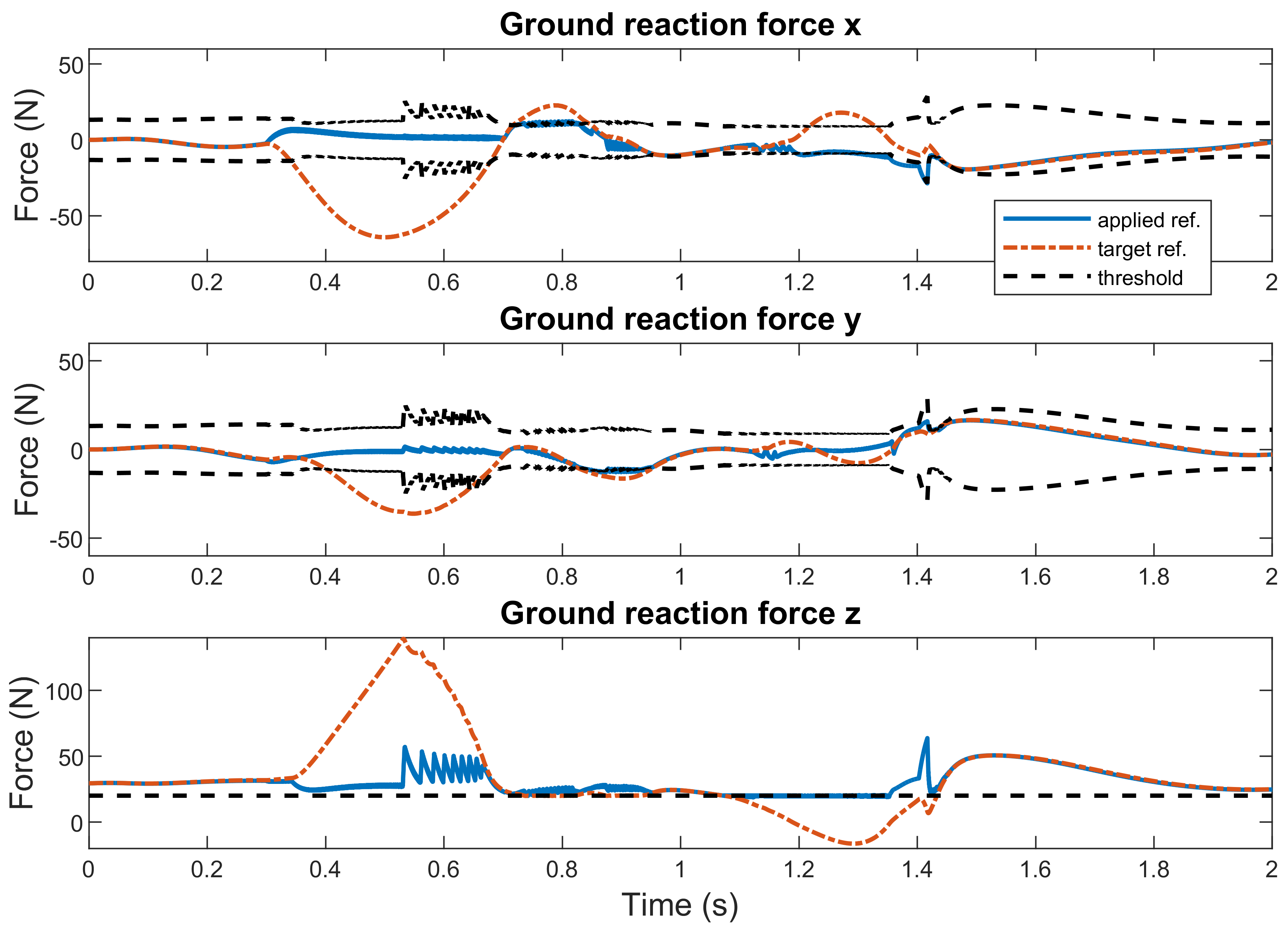}
    \caption{The simulated GRF in the VLIP model where the no-slip constraints have been successfully satisfied. These forces are the estimated GRF using the model shown in \eqref{eq:ground_forces}. Also, illustrates the difference between the GRF using target versus manipulated references.}
    \label{fig:plot_vlip_grf}
\end{figure}

\begin{figure}[t]
    \centering
    \includegraphics[width=1\linewidth]{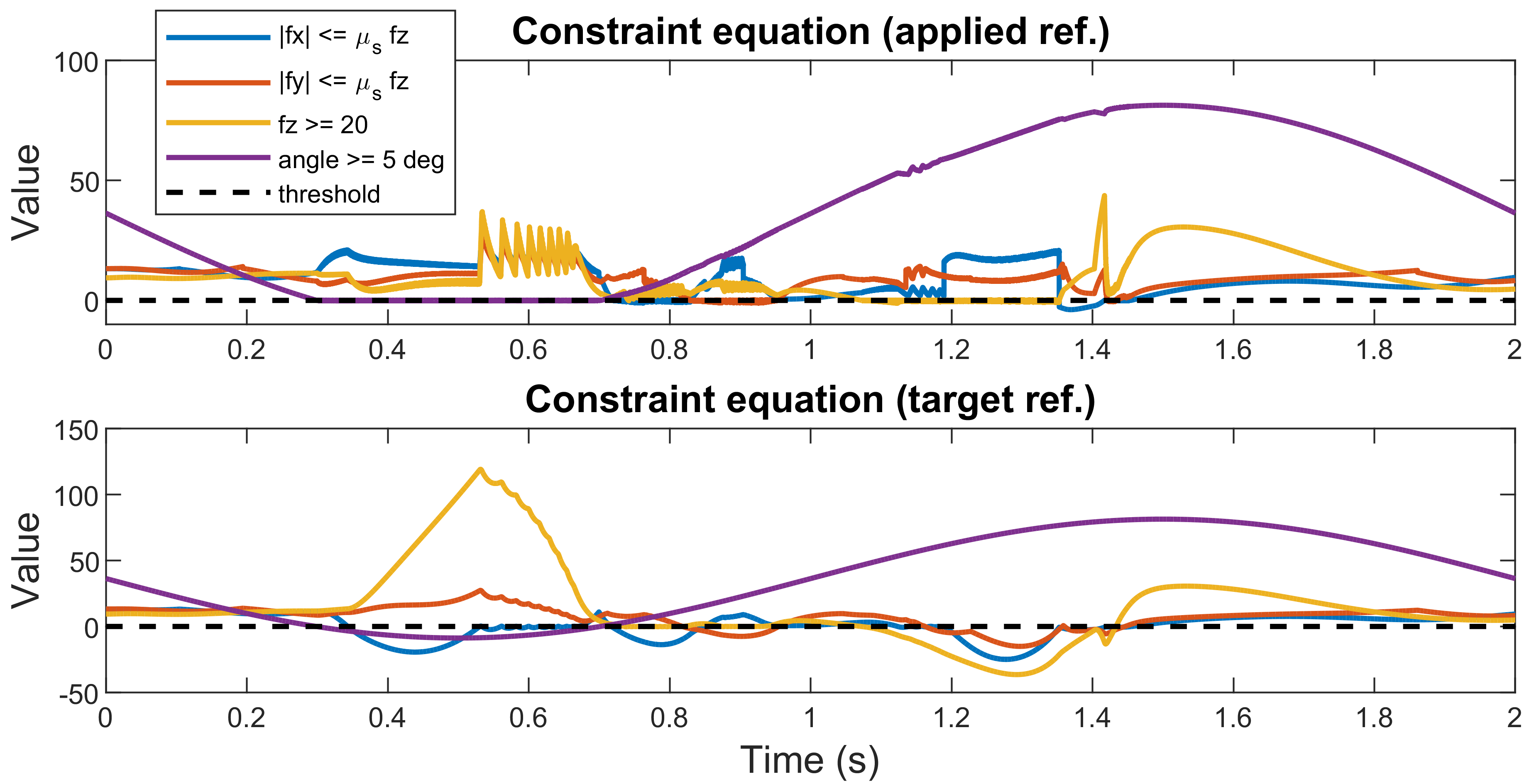}
    \caption{Illustrates the simulated constraints when the applied and target references are used in the VLIP model. When the target references are used instead of the manipulated trajectories, almost all constraints are violated.}
    \label{fig:plot_vlip_constraints}
\end{figure}

This section outlines the simulation setup and results of implementing the ERG in the VLIP model and the 3D Harpy model. The simulation using the VLIP model is done to show that the ERG is capable of manipulating the reference to enforce the specified constraints, which is then will be utilized in the 3D Harpy model.

\subsection{VLIP Model}

The ROM defined in \eqref{eq:harpy_rom} can be represented by the states $\bm q_{vlip} = [\theta, \phi, l]^\top$ instead of the body CoM position, where $\theta$ is the pendulum angle from vertical, $\phi$ is the pendulum heading, and $l$ is the pendulum length. A simulation of this ROM is done where we applied the ERG algorithm shown in Section \ref{sec:erg} to apply some constraints on the GRF and state trajectory. The target states references are defined as follows:
\begin{equation}
    \bm q_{vlip}(t) = \begin{bmatrix}
        0.45 + 0.45\sin(\pi t - \pi) \\
        - 1.5\sin(\pi t - \pi) \\
        0.4 + 0.1 \cos(\pi t - \pi)
    \end{bmatrix},
\label{eq:vlip_ref}
\end{equation}
and $ \dot{\bm q}_{vlip} = 0$. The following constraints are enforced:
\begin{equation}
\begin{aligned}
    \mu_s u_{g,z} - |u_{g,x}|  &\geq 0, \qquad & 
    u_{g,z} - 20 &\geq 0 \\
    \mu_s u_{g,z} - |u_{g,y}|  &\geq 0, & 
    \theta - 5^\circ &\geq 0,
\end{aligned}
\label{eq:vlip_constraints}
\end{equation}
where $\mu_s = 0.45$ and $\bm u_g$ is defined in \eqref{eq:ground_forces} and can be derived as a function of state reference using the control law in \eqref{eq:thruster_feedback_law}. The ERG is implemented in the controller \eqref{eq:thruster_feedback_law} by using the applied reference $\bm x_w$ instead of $\bm x_r$, where the ERG will drive it to be as close as $\bm x_r$ as possible while satisfying the constraints in \eqref{eq:vlip_constraints}. The simulation was run using the controller gains $k_p = 660$ and $k_d = 60$ for all the P and D gains respectively. Additionally, we initialized the simulation with $\bm x_w = \bm x_r$ and used the following ERG update rates: $\alpha_r = 1$, $\alpha_t = 5$, and $\alpha_n = 2$.

Figures \ref{fig:plot_vlip_states} to \ref{fig:plot_vlip_constraints} show the simulation result of the ERG application to the VLIP model. Figure~\ref{fig:plot_vlip_states} shows that the applied reference is significantly different than the target reference at around $t = 0.5$ s, which is done to avoid constraint violation. Figures \ref{fig:plot_vlip_grf} and \ref{fig:plot_vlip_constraints} show the comparison of the GRF and constraint equations between using the manipulated vs target references. The simulation result shows that most of the constraints are violated when using target reference while the manipulated reference keeps the constraint equation values above zero. There is a slight constraint violation near $t = 1.4$ s using the applied reference in Fig.~\ref{fig:plot_vlip_constraints}, which is quickly pushed into the specified threshold value by the $\bm v_n$ update law in \eqref{eq:wd_update_law}. This indicates that the ERG has successfully tracked the target reference \eqref{eq:vlip_ref} as closely as possible while satisfying the constraint equations \eqref{eq:vlip_constraints}. 

\subsection{Full-Dynamics of Harpy}

\begin{figure}[t]
    \centering
    \includegraphics[width=1\linewidth]{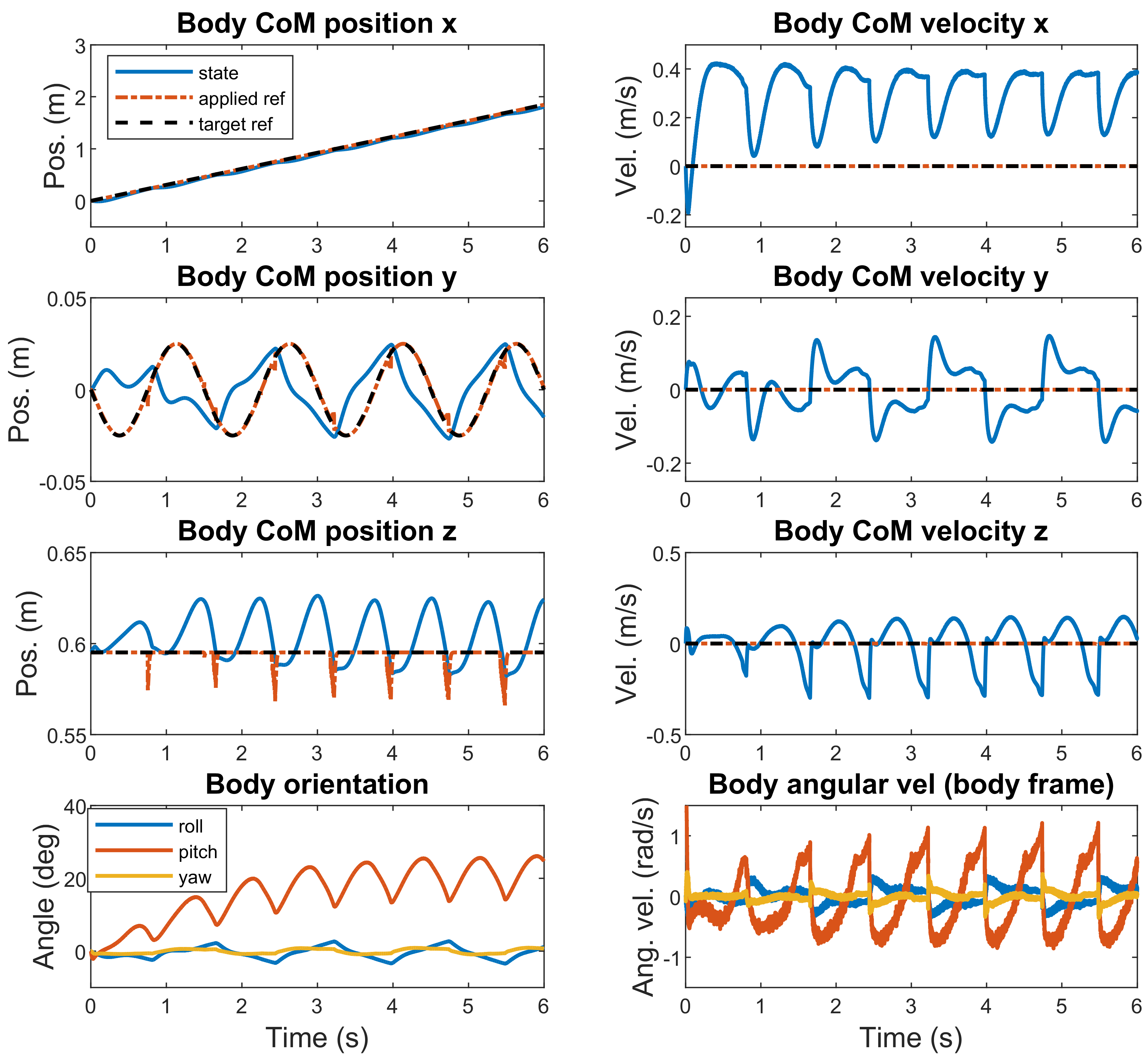}
    \caption{The evolution of the 3D-model states with the target and applied references.}
    \label{fig:erg_data_states}
\end{figure}

\begin{figure}[t]
    \centering
    \includegraphics[width=1\linewidth]{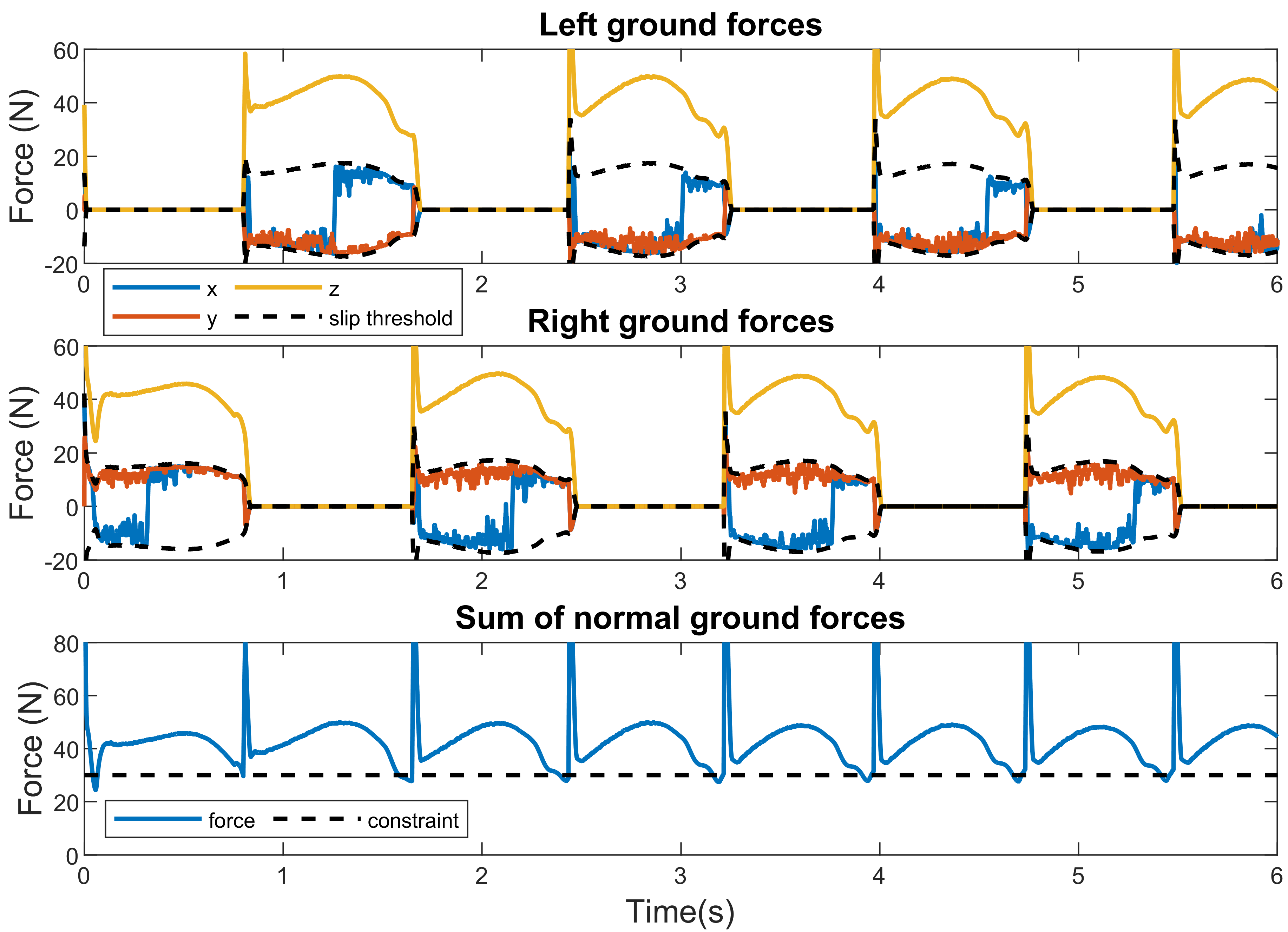}
    \caption{Illustrates the GRF in the 3D model after applying ERG which also shows the satisfaction of the no-slip constraints.}
    \label{fig:erg_data_ground}
\end{figure}

\begin{figure}[t]
    \centering
    \includegraphics[width=1\linewidth]{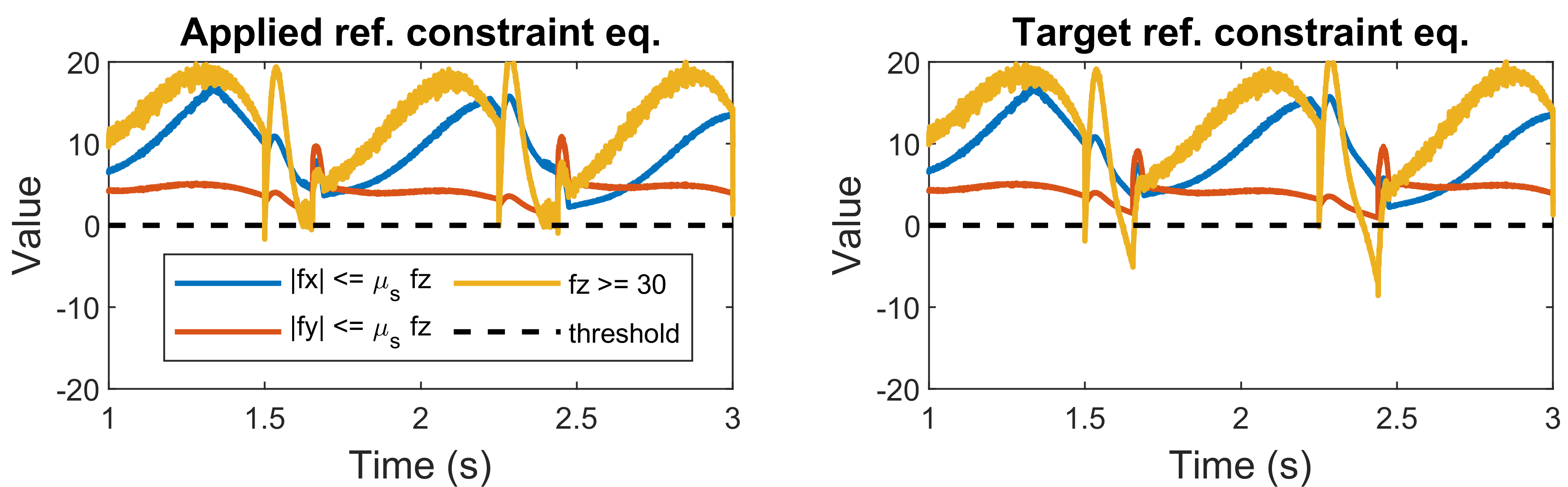}
    \caption{Illustrates the evolution of the constraints when applied ($\bm x_w$) and target ($\bm x_r$) references are employed. Note that when the target trajectories are employed the normal force constraints $u_{g,z}$ are violated.}
    \label{fig:erg_data_constraint}
\end{figure}

We must first find a stable walking gait for the robot in order to simulate the 3D Harpy model. This gait is found from a simulation where the frontal dynamics is ignored using an optimization technique and simple 4th order Bezier foot-end trajectories with a gait period of 0.75 s. This gait is not stable when implemented in the full 3D model, so the appropriate thruster forces defined in \eqref{eq:thruster_controller_full} are applied to stabilize the gait's frontal dynamics which allows the robot to walk stably using the 2D gait. Additionally, the controller for the 3D model can be calculated by using the ROM and ERG to satisfy the ground friction constraints to prevent slips. Currently $u_r$ is not used in the full model due to the potential clash with the foot end trajectories designed from this optimization. In exchange, we set $\bm Y = 0$ in \eqref{eq:thruster_feedback_law} to allow tracking about the pendulum's radial axis using the thrusters.

The ERG is implemented by estimating the GRF using the ROM in \eqref{eq:harpy_rom_summary} as the robot walks. The same GRF constraints as in \eqref{eq:vlip_constraints} are used in this simulation (sans the angle constraint), which can be derived using the reduced order states $\bm x = [\bm c; \dot{\bm c}]$, the reference states $\bm x_r = [\bm c_t ; \dot{\bm c}_t]$, and the center of pressure $\bm u$. The ground friction parameters used in the simulation are $\mu_s = 0.25$ and $\mu_c  = 0.225$ which makes the robot very prone to slipping. The target trajectory for this robot is simply a constant forward speed of 0.3 m/s, a sinusoidal lateral position with amplitude of 0.025 m and period of 1.5 s, and a constant height of 0.6 m. The controller proportional and derivative gains are set to be 400 and 40 respectively, and the following ERG convergence rates are used: $\alpha_r = 10$, $\alpha_t = 15$, and $\alpha_r = 20$.




The simulation results can be seen in Fig. \ref{fig:erg_data_states} which shows the target state references $\bm x_r$, applied references $\bm x_w$, and the body center of mass position states $\bm x$. The difference between $\bm x_w$ and $\bm x_r$ indicates that the ERG has modified $\bm x_w$ such that the resulting closed loop GRF followed the specified constraints. Figures \ref{fig:erg_data_ground} and \ref{fig:erg_data_constraint} show the ground forces and the constraints equations respectively. As shown in Fig. \ref{fig:erg_data_ground}, the robot avoids slipping by having containing the ground friction forces within the upper and lower bounds defined by the constraint equations $|u_{g,x}| \leq \mu_s u_{g,z}$ and $|u_{g,y}| \leq \mu_s u_{g,z}$. The constraint equations of both the applied reference and target reference ($\bm h_w$ and $\bm h_r$) indicates that the normal force constraint is violated frequently. The applied reference used seems to have successfully pushed the constraint equation back to positive region as it drops into the negative region. However, there is a significant chattering which is very likely caused by the non-smooth transition between the negative and positive $\bm h_w$.

\section{Concluding Remarks and Future Works}

The enforcement of the contact force constraints using Reference Governors (RGs) has been successfully demonstrated. To do this, we employed the thruster-assisted model of our bipedal robot call \textit{Harpy}. This robot is under development at Northeastern University. We demonstrated that an RG can manipulate the joint trajectories when using pre-defined gait parameters would lead to the violation of the constraints. For this purpose, we proposed an algorithm. This algorithm has minimum computational overhead and can potentially be used as an alternative to the computationally expensive optimization-based schemes.

The application of thrusters in our model, which allowed us to have full control over the unilateral contact forces, can lead to interesting path planning and trajectory generation problems. These problems will be a major part of our future research works.

\IEEEtriggeratref{11} 
\bibliographystyle{IEEEtran}
\bibliography{references.bib}

\end{document}